\documentstyle[named,11pt]{article}    

\newcommand{\krterm}[1]{{\sf #1}}
\newcommand{\lingform}[1]{{\it #1}}
\newcommand{\term}[1]{{\it #1}}
\newcommand{\code}[1]{{\tt #1}}
\newcommand{\acronym}[1]{{\sc #1}}
\newcommand{\scare}[1]{`#1'}

\newcommand{\fcomment}[1]{{\it ; #1}}

\newcommand{\IDAS}{{\sc idas}}
\newcommand{\KLONE}{{\sc kl-one}}

\newcommand{\class}[1]{{\bf #1}}

\title{Automatic Generation of Technical Documentation}

\author{Ehud Reiter\thanks{Now at CoGenTex, Inc, 840 Hanshaw Rd, Ithaca, NY
14850 USA. Email is {\tt ehud@cogentex.com}.},
Chris Mellish\thanks{Email is {\tt c.mellish@ed.ac.uk}.},
and John Levine\thanks{Email is {\tt j.levine@ed.ac.uk}.}\\
Department of Artificial Intelligence\\
University of Edinburgh\\
Edinburgh {\sc eh1 1hn}\\
Scotland}

\date{ }

\begin{document}

\setlength{\baselineskip}{13pt}		

\bibliographystyle{named}

\maketitle

\begin{center}
{\large\bf Abstract}\\
\end{center}
\begin{quote}
Natural-language generation (NLG)
techniques can be used to automatically produce
technical documentation from a domain knowledge base and linguistic and
contextual models.  We discuss this application of NLG technology
from both a technical and a usefulness (costs and benefits) perspective.
This discussion is
based largely on our experiences with the \IDAS{} documentation-generation
project, and the
reactions various interested people from industry have had to \IDAS{}.
We hope that this summary
of our experiences with \IDAS{} and the lessons we have learned from it will
be beneficial for
other researchers who wish to build technical-documentation
generation systems.
\end{quote}

\vspace{1in}

This paper will appear in {\em Applied Artificial Intelligence}, volume 9
(1995).

\newpage

\section{Introduction}

Natural-language generation (NLG) is a young but growing research field,
whose goal is to build computer systems that can automatically
produce fluent texts in English, French, and other human languages.
To date NLG has mainly been of interest to academic researchers, but
fielded application systems based on this technology have recently begun
to appear (e.g., \cite{goldberg:ieee94}).

In this paper, we discuss one particular
application of NLG technology, automatic generation of technical documentation,
mostly from the perspective of
\scare{what can the technology do and what are its costs and benefits},
instead of \scare{how does it work}.  Our discussion is
general, but much of it is based on our experiences with the \IDAS{}
technical-documentation generation project, which we worked on from 1990 until
1993.  \IDAS{} was a partial but not a complete success,
and we will discuss its weaknesses as well as its strengths, and the lessons
we have learned both from building \IDAS{} and from the reactions
interested people from industry have had to the system; these lessons will
hopefully be of use in future attempts to build technical documentation
generation systems.

\term{Technical documentation} refers to the extensive
design, maintenance, and operations documents that must be supplied with
complex machinery; we are primarily interested here in documents that are
intended to be read by technicians and other experts, as opposed to
\scare{the man in the street} (e.g., aircraft maintenance manuals, not VCR
operations manuals).  Producing technical documentation is a very expensive
process, and furthermore one that is relatively unautomated compared to
other aspects of the design process; in other words, there is as yet
no equivalent
in the technical-documentation field of the Computer-Aided-Design (CAD)
tools that have significantly enhanced the productivity of design engineers.
One potential candidate for the \scare{CAD tool for technical authors} is
systems that use natural-language generation technology to automatically
produce technical documents from various data- and knowledge-bases (KBs), and
the potential effectiveness and appropriateness of such tools is the subject
of this paper.

\IDAS{}, the Intelligent Documentation Advisory System,
was a three year collaborative effort between the University of Edinburgh,
Racal Instruments Ltd, Racal Research Ltd, and Inference Europe
Ltd.  Its goal was to build an advanced on-line documentation system for
users of Racal ATEs (automatic test equipment), which could give ATE users
help messages that were tailored to the context and the user.  Part of the
project involved an attempt to build what might be called an
\scare{advanced canned-text system},
which used hypertext and object-oriented techniques
to make a \scare{conventional} canned-text help and on-line-documentation
system both more effective
(for users) and easier to create, update, and otherwise modify (for authors).
The rest of the project was more ambitious,
and attempted to automatically generate documentation from a
domain KB and contextual models, using NL generation techniques; this is the
part of the project that the group at Edinburgh was most involved with,
and the focus of this paper.  As of the time of writing, the NLG system
seems less likely to be incorporated into Racal and Inference products than
the advanced canned-text system, essentially because it does not offer
sufficient benefits to make its extra cost worthwhile.
This is largely because when we started the project, we had only a very
vague idea about what the actual costs and benefits of using
NLG in document-generation applications were, and therefore did not
emphasize the benefits that turned out to be most significant.
Thanks largely to the valuable comments and criticisms about \IDAS{}
that we have obtained from various interested people in industry,
we now have a much better idea of
potential NLG costs and benefits and their relative importance, and we hope
that our presentation here of the lessons we have learned from \IDAS{}
will help future researchers who are interested in building
technical-documentation generation systems.

The rest of this paper will go over these points in more detail.
Section~\ref{tech-doc} will examine the general idea of producing
technical documentation from a knowledge base with NL generation techniques,
including a summary of the costs and benefits of this approach.
Section~\ref{IDAS} presents a summary of \IDAS{}: what it does,
how it works, etc.
Section~\ref{eval} will present an evaluation of the \IDAS{} NLG-from-KB
system, including the reactions of our collaborators and other interested
potential users as well as a summary of our user trials.  Finally,
in Section~\ref{lessons} we will try to summarize the lessons we have learned
that we think are most important for future efforts to build technical
documentation generation systems.

\section{Technical Documentation}
\label{tech-doc}

\subsection{The Problem}

Complex machinery of necessity requires complex documentation, and
producing technical documentation is a time-consuming and expensive task
for many corporations.  Stories abound of, for example, engineers who spend
five hours documenting for every hour they spend designing, or of airplane
documentation sets which weigh more than the plane they document.
In many cases, technical
documents also must meet externally imposed writing
or content standards, be translated into several languages, and be
written for easy maintenance and updating; all of these factors make the
documentation yet more expensive and time-consuming to produce.

The problem of generating technical-documentation in a cost-effective manner
is becoming even more critical because advanced Computer-Aided-Design (CAD)
tools are reducing the time required to design objects, but no equivalent tools
have been developed to reduce the time required to document designs.
Tools that could reduce document-creation time in a similar fashion to the
way CAD tools have reduced design-creation times would be of tremendous
value to numerous organizations, and also a boon to engineers (many of whom
find designing a more interesting and enjoyable task than documenting!).

\subsection{Using NL Generation}
\label{benefits}

Much of the information presented in technical documentation is
already present
in
machine-readable form in CAD systems, component-description databases,
knowledge bases created for expert-system applications, etc.
This suggests attempting to automatically create at least a portion of the
relevant technical document from this data, using natural-language
generation techniques.
The NLG approach has several potential advantages,
which are described below.

\subsubsection{Reduced Cost to Generate and Maintain Documentation}
\label{reduced cost}

If most of the information required to generate the documentation is
already present in a database or knowledge base of some kind,
then the NLG approach can reduce the time and effort required
to produce documentation.
This should be the case even if the system's output needs to
be post-edited by a human, or if additional
information needs to be entered into
the system to support document generation, provided the amount of post-editing
and extra information required is not over-large.

Cost savings may be even more significant for document maintenance than
initial document creation.  As with software,
maintaining and updating a document can be more costly than initially
creating it; the problem here is not so much fixing spelling and grammatical
mistakes in a document, but rather keeping a document up-to-date when the
machine it describes is upgraded or released in a new configuration.
In many cases such upgrades and new
configurations can be represented by very simple changes to a design database
that describes the machine,
and the NLG approach allows
the new documentation to be produced automatically once the design
has been upgraded in this manner.

For example, if a machine is upgraded by installing a higher-capacity
power supply, the NLG approach allows the specifications of the new power
supply to be loaded
into a single well-defined place in the domain KB, from which they
will be automatically propagated to all relevant documentation texts.
In many cases this can be significantly easier than manually making such
changes, particularly if references to the power supply are scattered
throughout the documentation set.

\subsubsection{Guaranteed Consistency Between Documentation and Design}
\label{guaranteed consistency}

Current practice often requires engineers to instantiate their designs twice:
once in a design database (e.g., CAD system), which may be used to drive
automated Computer-Aided-Manufacturing (CAM) equipment;
and once in a human-readable document,
where maintainers and users can learn about it.
This duplicate instantiation is of course expensive, but perhaps even
more significant is the danger of inconsistencies.
If the design described in the document is not exactly the same as the design
entered in the design database, the user may misuse the machine, and the
manufacturing company will be held legally liable if its documentation
was not correct.  The NLG approach allows the designer to
instantiate his or her design once, into a design database or knowledge base
that is augmented to also
represent the extra data needed for document generation;
this greatly reduces the possibility of errors
due to inconsistency between the machine- and human- readable representations
(as well as potentially reducing effort,
since designers only have to instantiate their designs once instead of twice).

As with the reduced cost benefit (Section~\ref{reduced cost}), guaranteed
consistency can be especially important for document maintenance.
If a machine is upgraded or released in a new configuration, it is very
easy for the documenter to forget to make some of the necessary changes
in the documentation, especially if the current document maintainer is not
the original document author; the NLG approach can
significantly reduce the risk of this eventuality.

\subsubsection{Guaranteed Conformance to Standards}
\label{standards}

Many documents are required to obey writing or content standards.
Writing standards are usually intended to ensure that the language used
in a document is unambiguous and easily comprehensible, especially for
non-native English speakers.  Examples of such standards include AECMA
Simplified English \cite{simplified-english:book} and Perkins Approved
Clear English \cite{perkins:pace}.
AECMA Simplified English, for example,
\begin{itemize}
\item Imposes a fixed unambiguous lexicon.
\item Prohibits potentially confusing syntactic constructs, such as gerunds
or complex tenses.
\item Imposes general stylistic guidelines, such as requesting that sentences
be kept under 20 words if possible.
\end{itemize}

Content standards, such as the UK Army Equipment Support Publication
rules, and the US Defense Department 2167A standard
for software documentation,
specify what information must be included in various documents (e.g., required
maintenance procedures and safety information).
Content standards are often less precise than
writing standards, which can make them harder to automate.

NLG systems can be set up to automatically enforce the rules of any
given writing standard, by programming an appropriate grammar and lexicon
into the system.  NLG systems can also be set up to obey content standards
if the relevant information is available in the knowledge base,
and print a warning message
if it is not, provided that the standard is precise enough to be
computationally interpretable.

\subsubsection{Multilinguality}

If the relevant domain and contextual models are language-independent, then
the NLG system can be modified to produce texts in multiple output languages;
NLG systems with multilingual output have in fact been built since the
NLG field began \cite{inlgw92:multilingual-panel}.
Producing multilingual output is not a trivial technical problem, but it
is perhaps less complex for technical documentation than for other kinds of
text, since
\begin{itemize}
\item Complex and difficult-to-translate syntax, lexemes, tenses, etc.
are prohibited
by most technical documentation writing standards (Section~\ref{standards}),
and hence the system does not have to worry about correctly using such
complex linguistic constructs in multiple languages.
\item Achieving complex pragmatic and stylistic effects (e.g., making the
reader laugh, or indirectly informing them of a fictional
character's mental state)
is not generally a goal of technical documentation; these are some of the
most difficult things to get right in multilingual texts.
\end{itemize}

Multilingual output reduces document translation costs, but it probably
will not eliminate it completely, since it is likely that human quality
assurance and post-editing will still need to be performed for texts in all
output languages.

\subsubsection{Tailoring}
\label{tailoring}

The NLG approach allows a documentation text to be dynamically
tailored to the context (e.g., the user's task, the user's expertise level,
and the discourse history).  Among the many kinds of tailoring that
have been discussed in the literature are:
\begin{itemize}
\item Tailoring rhetorical \cite{paris:cl} and syntactic
\cite{bateman-paris:ijcai89} structures according to a user's expertise.
\item Choosing different lexical units (words) depending on the user's
vocabulary and background knowledge \cite{reiter:ci}.
\item Generating helpful responses that communicate the information the
user needs to execute his or her current plan \cite{allen-perrault:aij}.
\item Choosing appropriate referring expressions for the current
environment and discourse context (e.g., \cite{reiter-dale:coling92}).
\end{itemize}
The research literature on this topic is extensive, and the above list is
by no means complete.

\subsubsection{Multimodality}

Information can be communicated to (and from) the user with graphics
as well as text; ideally, a document generation and presentation system
should be able to interact with the user in whatever modality is most
suitable for the task at hand.  It is useful to distinguish between three
kinds of multimodality:
\begin{description}

\item[Visual Formatting:]
Text can be much more effective if it is presented with appropriate
visual formatting devices, such as bulletization, font changes, indentation,
etc.  An NLG system can produce visually-formatted text by treating such
formatting devices as an additional \scare{resource} that can be used
to communicate and structure information
(e.g., see \cite{hovy-arens:aaai91}).

\item[Hypertext Input:]
Text can be generated with hypertext-like links that allow a user to
issue clarification, elaboration, and other kinds of followup questions
simply by clicking his or her mouse on an appropriate word
\cite{alfresco:aai93,idas:applied-acl,moore:aaai}.
This is obviously only useful in an on-line system.

\item[Graphics Output:]
Much research has been done on generating diagrams (and other graphical
presentations of data) and text from a single domain knowledge base,
e.g., \cite{feiner-mckeown:aaai90,wip:aij}.
Graphics output requires a different low-level
\scare{realisation/rendering} module than text output,
but in some cases high-level content-oriented modules can be be used
for both text and graphics output \cite{wip:aij}.

\end{description}
%
NLG techniques can be adopted to the problem of producing appropriate
visual formatting and hypertext links, and to determining the content
(although not the layout) of associated diagrams.
When doing so, many of the advantages mentioned in the previous
sections also apply to these multimedia \scare{extensions}.  For example,
AECMA Simplified English (Section~\ref{standards}) has standards for
the use of visual formatting, which an NLG system can ensure are obeyed;
hypertext links can be automatically updated if the text they point to is
changed; and generated diagrams can be modified according to the user's
goals \cite{roth:chapter91}.

With hypertext in particular, it is also possible that automatic
generation of hypertext links may produce a more consistent and therefore
easier to navigate hypertext network.  This currently remains an interesting
but unproven hypothesis; more research needs to be done on the acceptability
of automatically generated hypertext networks.

It is also worth noting that NLG systems will be much more useful in
practice if they can include visual formatting, hypertext links, and
associated diagrams in their output.  An NLG system that has no
multimodal abilities may only be useful in a limited number of real-world
document-generation applications.

\subsection{Costs of NLG}
\label{costs}

Against the potential benefits must be weighed the costs of the NLG
approach; NLG will, of course, only be worth using in real applications
if its benefits outweigh its costs.

\subsubsection{Increased CAD/KB creation time}
\label{cad/kb creation time}

In general design and other databases do not contain
all the information needed to produce the relevant documentation, which means
the designer/engineer will need to enter additional information into the design
database or KR
system when creating his or her design,
in order to give the NLG system sufficient
information.  This extra information can, however, be used for many other
purposes as well as document generation, including consistency,
correctness, and completeness checks on the design.  The cost of creating
an appropriate model of a system in a CAD or KR
framework should thus be evaluated in light of all the benefits it can
potentially bring, not just document generation.

\subsubsection{Fixed overhead for KB creation}
\label{fixed overhead}

The NLG system will also require
knowledge bases that describe the sublanguage
used in the documentation (which is often specified in a writing standard),
and user and contextual models (if tailoring is being done).
The per-application cost
of building these knowledge bases will be decreased if the KBs
can be shared among several applications, which is certainly possible
to some degree (e.g., a grammar and lexicon for the AECMA Simplified
English sublanguage can probably be used for most documents about
aerospace systems).

\subsubsection{Quality Assurance}
\label{quality assurance}

Many organizations require documents to pass through a Quality Assurance (QA)
procedure, which usually means being checked and perhaps edited by a
separate group of people (this last procedure is sometimes called
\term{post-editing}).
It seems likely that computer-generated documents
will also have to pass through this QA process, at least until users
are confident that generated documents are both linguistically
correct and a faithful rendition of the relevant knowledge base or database.
Such checking and post-editing can cost significant amounts of money
(e.g., see the costs reported in \cite{taum-aviation:cl} for post-editing
in a machine translation project).

\subsubsection{Computation Time}
\label{computation time}

A certain amount of computer time will obviously be required to generate text
using NLG techniques.  While the monetary cost of computer time is
fairly low (and getting lower), NLG systems must satisfy response-time
constraints.  In particular,
interactive systems must be able to generate text within a
few seconds in order to be useful.  The response-time constraints on offline
(batch) generation are looser, but they exist;
a batch system that required several days to generate a document, for example,
would probably be considered to be of limited usefulness.

\subsection{Related Approaches}

There is some overlap between generating documentation from a KB with NLG
techniques, and using Knowledge-Based Machine Translation (KBMT)
\cite{kbmt:book} techniques to translate documents.
KBMT systems take an input document
written in one language, process that document with an NL understanding
system to produce an \scare{interlingua representation}
that essentially contains
the same information as a pure NLG system would hope to extract from its
various knowledge and databases, and then use NLG to produce an output text
from the \scare{interlingua}.
{}From a technical perspective,
the main difference between the KBMT and pure NLG
approaches is that the former expects its input data to be expressed as
NL text (in a different language), while the latter expects it to be present
as design information in a database of some kind.
{}From an applications
perspective, it is worth noting that
KBMT is generally viewed only as an aid
to document {\em translation}, while NLG can be used to improve productivity
throughout the document creation process.

\section{IDAS}
\label{IDAS}

\subsection{Goals}

The \IDAS{} project was a collaboration between the University of Edinburgh,
Racal
Instruments Ltd, Racal Research Ltd, and Inference Europe Ltd.
Its goal was to
build a better on-line documentation system for Racal ATEs (automatic
test equipment); these are complex machines that are used to test potentially
faulty circuit boards and determine if they are in fact malfunctioning.
\IDAS{} was intended to produce
short on-line help messages (as opposed to complete
paper documents) for three kinds of ATE users --- operators, maintenance
technicians, and programmers.  Two systems were built:
\begin{itemize}
\item A hypertext documentation system which mainly relied on canned texts,
but which used a domain KB to enhance the effectiveness of the system
in various ways
(somewhat similar to what \cite{hayes-pepper:hypertext} proposed, but
did not implement).  The system
used object-oriented techniques to make the documents easier to update
and otherwise modify.
\item An NLG-based system which in addition to the above, attempted to
generate the hypertext nodes (both text and links) from a domain KB and
various contextual models.
\end{itemize}
Our group at the University of Edinburgh
was primarily concerned with the second of these systems, and
this is the one this paper focuses on.

In relation to the benefits described in Section~\ref{benefits},
the initial goals of the \IDAS{} NLG system could be characterized as follows:

\begin{description}
\item[Reduced cost:]
The main interest was in reducing document maintenance costs.
ATE designs were not available in machine-readable databases, which meant
that special KBs would need to be constructed for the NLG system,
and this would probably cost more than simply directly authoring the
documentation.  The hope, though, was that once an NLG KB had been built,
changes to reflect new ATE configurations, or upgraded ATE components,
could be made easily in the KB, and this would reduce document maintenance
costs (which are high, since ATEs are sold in many different
configurations, and are continually being upgraded to utilize the most
up-to-date components).
\item[Guaranteed consistency:]
This was a significant goal, especially for document maintenance.
It was not possible to ensure consistency between the document and the design
(since the design was not present in a CAD system), but it was hoped to
increase the likelihood of consistency by making it more straightforward
to update the documentation.
With \IDAS{},
the designer or technical author could update the documentation to reflect
modified hardware simply by changing the KB to reflect the changes in the
hardware, and all necessary documentation changes would then be made
automatically.
\item[Standards:] Not emphasized.
\item[Multilinguality:] Not emphasized.
\item[Tailoring:] This was also important; a primary goal of the system
was to be able to tailor its output to at least the three classes of users
mentioned above (operators, maintenance technicians. programmers).
\item[Multimodality:] Hypertext was central to the project; some importance
was also attached to being able to use canned graphics.
\end{description}

With regard to cost, the main concern was to reduce the cost of authoring
the domain KB as much as possible; this was especially critical because it
was not possible to extract any information from existing databases.
Less emphasis was placed on reducing the cost of constructing fixed KBs,
since it was felt that this cost could be amortised over several projects
if \IDAS{} was successful.  The main computation constraint was that
response texts should be generated in an acceptable time for an interactive
system, i.e., a few seconds.
Quality assurance was not originally regarded
as a significant cost, although in retrospect it did have an impact,
especially when considering the amount of tailoring that was desirable.

Some \scare{intermediate} techniques were developed which attempted to reduce
domain-KB authoring costs at the expense of making some relatively
unimportant benefits (e.g., multilingual generation) more difficult to
achieve; these are discussed in Section~\ref{intermediate techniques}.

\subsection{The System}

\subsubsection{Input}

The prototype \IDAS{} NLG system built by Edinburgh is described
in \cite{idas:applied-acl}.  The system's input is a
\scare{question space} point that specifies five parameters
\begin{description}
\item[Basic question:]
The basic system supported seven questions:
What-is-it, Where-is-it, What-are-its-parts,
What-are-its-specs, What-is-its-purpose, What-does-it-connect-to,
and How-do-I-perform-the-current-task.  This list was modified
for some of the non-ATE knowledge bases.
\item[Component:]
The knowledge base contained a Part-Of component hierarchy of the target
machine (the ATE in the main \IDAS{} application),
and queries could be issued for components at any
level (from the ATE as a whole down to individual switches and levers).
\item[User-Task:]
The user-task model told \IDAS{} (in very rough terms) what kind of task
the user was performing. The tasks were represented in an
IS-A taxonomy.
\item[User-Expertise:]
The user-expertise model told the system how much the user knew about the
domain, and what some of his or her stylistic preferences were.
The former included
what technical vocabulary the user knew and what actions he or she 
could perform;
the latter included, for example, whether contractions should be used
(e.g. \lingform{it's} vs.\ \lingform{it is}).
\item[Discourse:]
This told the system what objects were salient and hence could be referred
to by simple noun phrases; this follows a much simplified version of the
discourse model proposed by \cite{grosz-sidner:cl}.
\end{description}

For example, the question space point $\langle$What-is-it,
\class{DC-Power-Supply-23}, Operations, Skilled,
\{\class{VXI-Chassis-36}, \class{DC-Power-Supply-23}\}$\rangle$
represents the query \lingform{What is the DC Power Supply?}
when asked by a user of Skilled
expertise who is engaged in an Operations task with the discourse context
containing the objects \class{VXI-Chassis-36}
and \class{DC-Power-Supply-23}.
The NL Generation component would in this case produce the response:

\begin{quote}
\lingform{It is a black Elgar AT-8000 DC power supply.}
\end{quote}

More example \IDAS{} outputs, including ones that show the effect of changing
the user-task or user-expertise models, are shown in Figure~\ref{example-dump},
and described in Section~\ref{example}.

\subsubsection{Knowledge Base}
\label{knowledge base}

A \KLONE{} \cite{klone:bs} type knowledge representation language called I1
was used as \IDAS{}'s knowledge
representation system.  I1 included support for IS-A and Part-Of hierarchies,
default attribute inheritance (along the IS-A hierarchy), and
automatic classification of new classes into the correct position in the
IS-A taxonomy.  In addition to the basic KR support functions, I1 also
included a graphical browser that could be used to examine the knowledge base.

I1 proved surprisingly powerful and versatile; \IDAS{} used it to represent
many kinds of information, including:
\begin{itemize}
\item domain knowledge;
\item grammatical rules;
\item the lexicon;
\item user task and expertise models;
\item content-determination rules.
\end{itemize}
\IDAS{} also used I1's classification
and inheritance mechanisms to perform most of the reasoning needed to
generate text \cite{idas:acl}.  The use of a single KR system for so many kinds
of knowledge and so many kinds of reasoning is perhaps the most
theoretically interesting feature of \IDAS{}.

{}From a practical perspective, the use of an object-oriented KR system
that supported taxonomies and inheritance made it significantly easier
to create the necessary knowledge bases.  For example, the procedure for
removing a circuit board from a VXI chassis (a type of backplane used in
ATE systems) was only specified once, at the level \krterm{VXI-chassis-board},
and then inherited by all the specific boards (digital multimeter, counter
timer, etc).  Inheritance was also used within the linguistic knowledge
bases; the definition of the grammatical rule for \krterm{Imperative-Sentence},
for example, was relatively short because it could inherit most of the
necessary information from its parent class \krterm{Sentence}.  It is
unclear to what extent the presence of a default inheritance system added
to the theoretical expressive power of I1, but it certainly proved to be
a significant convenience to KB authors.

Most of the knowledge bases \IDAS{} was used with were created by hand;
our experience showed that a domain KB for a machine with 50 subcomponents
could be created in a few weeks by someone familiar with the system and
knowledgeable in AI techniques (Section~\ref{bike kb}).
Two of our industrial collaborators, Inference Europe and Racal Instruments,
developed a graphically-oriented KB authoring tool that
could be used by people who were less familiar with AI techniques.
This tool, for example, used the terms
\term{parts stores} and \term{family trees}, which are standard Racal
terminology, instead of \term{IS-A taxonomy} and \term{Part-Of hierarchy}
(AI terminology); it also attempted to use some of Racal's standard
presentation techniques
for describing \scare{parts stores} and \scare{family trees}.

Unfortunately, the Inference/Racal authoring tool was not developed until
fairly late in the project, and therefore it has not yet been used to construct
a non-trivial \IDAS{} knowledge base.  There was also a feeling that the
authoring tool would be more useful if it could be used to build up a general
purpose design description that could be used for other tasks as well as
document generation.  As this paper is being written, further research is
being considered to extend the authoring tool in this manner,
and to investigate how it would best fit into the Racal design and
documentation environment.

\subsubsection{Operations}
\label{operations}

\IDAS{}, like many other applied NLG systems \cite{reiter:arch},
generates texts in three stages:

\begin{description}

\item[Content Determination:]
The basic-question, component, and user-task
components of the question-space tuple are used to pick a
{\em content-determination rule}.  This rule specifies:
\begin{itemize}
\item The basic structure of the response, i.e., the \term{schema} used to
build it (see \cite{mckeown:book}, although our schemas have a somewhat
different structure than McKeown's).
\item The information from the knowledge-base that will be included in the
response text.
\item Hypertext followup buttons that will be displayed at the bottom of the
response text.  The idea is that information that is immediately
relevant should be presented in the response text, while information that may
possibly be relevant should be accessible by clicking on a followup button.
\end{itemize}
We used a rule-based content-determination system in \IDAS{}, because
we believed rules would
be relatively easy for domain experts to create \cite{idas:ijcai}.
The rule-based system was also very fast, which was important in ensuring
acceptable response times.

\item[Sentence Planning:]\footnote{In
some earlier papers, we referred to this process as `text planning' instead
of `sentence planning'.  We use the term `sentence planning' in this
paper because we believe it is more consistent with the terminology used
by other researchers \cite{reiter:arch}.}
An SPL \cite{kasper:darpa-workshop} expression (i.e., a semantic form) is
constructed from the output of the content-determination system.
This process is sensitive to the user-expertise and discourse components
of the question-space tuple, and involves, in particular:
\begin{itemize}
\item \term{aggregation} \cite{dalianis-hovy:nlgw93}, ie
deciding how many sentences to use, and which information should be
conveyed by each sentence.  This is currently done fairly simplistically;
a more complex aggregation and optimisation module was developed as an
MSc project \cite{pake:msc}, but it was not reliable enough to be used
in the main version of the system.
\item {\em Generating referring expressions}.
Pronouns were generated by a
simplified version of the centering algorithm \cite{gjw:acl};
definite noun phrases were generated with the algorithm
described in \cite{reiter-dale:coling92}.
\item {\em Choosing appropriate open-class lexical items (words)}.
This was
based on the ideas presented in \cite{reiter:ci}, and involved, for example,
trying to use \term{basic-level} terms \cite{rosch:chapter} whenever possible.
\end{itemize}

\item[Surface Realization:] The SPL term is converted into a surface form,
i.e., a set of words with formatting and hypertext annotations.
This process involves:
\begin{itemize}
\item {\em Syntactic processing}.
The \IDAS{} grammar is represented as a series
of I1 classes, and classification is used to apply the grammar to the SPL
produced by the sentence planner \cite{idas:acl,mellish:chapter}.  The \IDAS{}
grammar is small when compared to, for example,
ISI's \acronym{nigel} \cite{nigel:acl}
grammar or Elhadad's \acronym{surge} grammar \cite{elhadad:thesis},
but it is adequate
for \IDAS{}'s needs (remember that writing standards for technical
documentation generally prohibit complex syntactic structures in any case).
\item {\em Morphology}.
Morphological processing in \IDAS{} is again done with classification;
some of the specific rules are taken from \cite{ritchie:book}.
A morphological processor for Romanian (which is much more complex in
morphological terms than English) was also built within the \IDAS{}/I1
framework \cite{cristea:project}.
\item {\em Post-processing}.
This module handles capitalizing sentence-initial words, inserting the right
spacing around punctuation (e.g., \lingform{My dog (Spotty) is here},
not \lingform{My dog( Spotty )is here}), and other such details
of the written form of English.
\end{itemize}

\end{description}

\IDAS{}'s NL generation system was only designed
to be able to generate small pieces of text (a few sentences, a paragraph
at most).  This was because \IDAS{}'s hypertext system enabled users to
dynamically select the paragraphs they wish to read, i.e., perform their
own high-level text planning \cite{levine:nlg-workshop},
thereby eliminating the need for the generation system to perform
such planning.

\begin{figure*}
\vspace{1in}
\caption{Example Screen Dump}
\label{example-dump}
\end{figure*}

\subsubsection{Example}
\label{example}

Figure~\ref{example-dump}
shows several complete \IDAS{} texts (including hypertext followup buttons).
The texts are shown in a simple hypertext display system developed at
Edinburgh; a more sophisticated hypertext delivery system was built by
our industrial collaborators.
The initial query was What-are-its-parts, asked about the
complete ATE by
a Skilled expertise person performing an Operations task; this produces
the text shown in Response 1.
The underlined part names (which are in fact
referring expressions) are all mousable, as is \lingform{ATE}
in the title question and the buttons on the bottom line.
Response 2 shows how the system would respond to the same query issues
under a Naive user-expertise model.  Note in particular that the components
described in Response 1 as \lingform{the DC power supply} and
\lingform{the mains control unit} are described in Response 2 as
\lingform{the black power supply} and \lingform{the silver power supply};
this is a consequence of the fact that Naive users are not expected to have
as rich a technical vocabulary as Skilled users.

Response 3 was produced by clicking on {\tt test head} in Response 1,
and selecting What-is-it from a pop-up menu of basic questions;
this response was generated using the same user-task,
user-expertise, and discourse-in-focus question-space components as
Response 1.
The \krterm{What-Operations-Rule} content rule used to generate Response 3
specifies that Where-is-it and How-do-I-Use-it should be added as hypertext
followups,
so {\tt WHERE} and {\tt USE} buttons are presented below the
text.
Other questions, e.g., What-are-its-parts, can be asked by clicking on
{\tt test head} in the title question, and selecting from the pop-up menu.
The {\tt MENU} button allows the user to change the contextual models
(user-task, user-expertise, etc).  Response 4 shows the response for the
same query under a Repair-Part task.  More information is
given in the response text (for example, a reorder part-number is included
in the Repair-Part response but not the Operations response),
and also more followup buttons are created
(e.g., {\tt SPECS} (specifications) are assumed to potentially be of interest
to a maintenance engineer, but not to an operator).

Response 5 was obtained by clicking on {\tt WHERE}; it answers
\lingform{Where is the test head?}.
Response 6 comes from clicking on the {\tt USE} button in
Response 3;
it is a response to \lingform{How do I use the test head?}.
In this response the
underlined nouns {\tt test head},
{\tt ITA mechanism}, and {\tt ITA}  are all linked to pop-up menus of
basic questions about these components, while the verbs {\tt unlock},
{\tt mount}, and {\tt lock} are all linked to How-do-I-perform
queries for the relevant action.  Clicking on {\tt unlock} produces
Response 7, which presents a step-by-step decomposition of the action
of unlocking the ITA mechanism.
Response 8 was obtained by clicking on {\tt lever} in Response 7,
and selecting What-is-it from the pop-up menu.

\begin{figure*}
\vspace{1in}
\caption{A Trace of Response 8 in Figure 1}
\label{example-trace}
\end{figure*}

Figure~\ref{example-trace} shows a trace of \IDAS{} generating Response 8
in Figure~\ref{example-dump}.
The initial query can be textually represented as
\lingform{What is the lever?}, but is represented internally as a
What-is-it query about the test-head's locking-lever under the context of
an Operations task undertaken by a Skilled user; the default discourse
context is used.

This input triggers the \krterm{What-Operations-Rule} content-determination
rule.  This specifies that text should be structured by the
\code{Identify-schema} (which basically means a single \lingform{is a} sentence
will be generated), with no bulletization being performed, and with
no abbreviations allowed
(e.g., \lingform{digital multimeter} would be used instead of \lingform{DMM}).
The text should directly inform the user that \krterm{Llever-test-head12} has
the property \code{(colour black)}, and Where-is-it should be presented
as a hypertext followup button.\footnote{What-Operations-Rule was
also used to generate Response 3 in Figure~\ref{example-dump}.
The difference between the information
conveyed in Responses 3 and 8, and the followup buttons shown, is due
to the presence/absence of
knowledge in the knowledge base; for example, the {\tt USE} button is
present in Response 3 but not in Response 8 because the
knowledge-base does not have information about how people use levers.}

This output is given to the sentence planner, which generates the SPL shown in
Figure~\ref{example-trace}.\footnote{Our version of SPL differs
from the original \acronym{penman} version \cite{penman-user-guide}.}
Note that the referring-expression module has decided
to pronominalize the subject (i.e., \code{:DOMAIN} filler), based on its
discourse model and centering rules, and that the lexical choice module
has decided to use the noun \lingform{locking lever} for
\krterm{Llever-test-head12}, instead of just \lingform{lever}.

The final stage is converting the SPL into a surface form, i.e., actual text;
the output of the surface realisation module is
\lingform{It is a black locking lever.}

Each of the responses in Figure~\ref{example-dump}
was produced in less than two seconds on a SUN IPX (Sparc 2) workstation,
measured from the initial click on a hypertext followup button to the
appearance of the response box on the user's screen.
Almost all \IDAS{} responses are in fact produced within two seconds,
and this seems acceptable to users.

\subsection{Intermediate Techniques}
\label{intermediate techniques}

Since not all of the benefits listed in Section~\ref{benefits} were deemed
important in \IDAS{}, we decided to search for generation techniques that
\scare{cheated} in certain ways and hence sacrificed some of the benefits
listed in Section~\ref{benefits}, but in return lowered some of the costs
listed in Section~\ref{costs}; if we could sacrifice benefits that were
unimportant in \IDAS{} and as a result decrease costs that were deemed
quite important, then this would make the system more useful and valuable.
\cite{idas:ijcai} describes our search for such \term{intermediate techniques}
in more detail; in this paper, we will just describe one such technique,
the use
of \term{hybrid action representations}, to give readers a feel for what
an intermediate technique consists of, and how it is motivated.

One of \IDAS{}'s tasks is to tell the user how to perform actions
(e.g., Response 6 in Figure~\ref{example-dump}).
\IDAS{} can perform this in a \scare{deep} manner by generating text from
a case-frame representation of the action to be performed; it is, however,
impractical to expect domain experts or technical authors,
who in general have minimal experience with AI techniques, to create such
case frames by hand.
Such authors find writing text easier than building case frames, so the ideal
solution would be to have the authors write text and then convert this text
into case frames with an NL understanding system.
Given the state of the art in NL understanding, it is
difficult to reliably and unambiguously translate arbitrary
texts into \IDAS{}'s
internal case-frame notation, but some processing can certainly be done.
This has led to the notion of \scare{hybrid} action representations which
mix proper knowledge-base structures with canned-text fragments; the former
represent pieces of the input text that the analysis system can confidently
analyze, while the latter are used for unanalyzable portions of the text.

More specifically, we support two hybrid representations in \IDAS{}:
{\em canned text with embedded KB references} (EKR), and
{\em case frames with textual case fillers} (TCF).  In the EKR representation,
references to machine components and other KB entities can be embedded in
a canned-text action representation; the generation system then generates
appropriate referring expressions for these references when it processes
the EKR form (this is somewhat similar to the system described
by \cite{springer:iaai}).
In the TCF representation, the \IDAS{}
case-frame representation is used, but case fillers are allowed to be
canned text; these are then inserted into the generated sentences in
appropriate positions.
Examples of these representations are:

\begin{description}
\item[Canned-text:]
\lingform{Remove any connections to the board}
\item[EKR:]
\lingform{Carefully slide \krterm{[Board21]} out along its guides}
\item[TCF:]
\krterm{REMOVE(actor=User, actee=Board21, source=Instrument-Rack1,
manner=``\lingform{gently}'')}
\item[Case-frame:]
\krterm{PUT(actor=User, actee=Board21, destination=Faulty-Board-Tray3)}
\end{description}

Along with our industrial collaborators and an MSc student,
we have developed authoring tools
that can produce EKR or TCF representations from textual input; one of
these tools also has some support for graphical authoring of actions
\cite{marshall:msc}.  Entering an EKR or TCF action specification with
one of these tools (or indeed by hand)
is usually quicker than manually building up a case-frame structure;
perhaps more importantly, it also requires less detailed knowledge of how
information is represented in \IDAS{} and I1.
The cost of using these
techniques is that some of the potential NLG benefits described in
Section~\ref{benefits} are lost.  In particular, {\em multilingual generation}
is impossible, and {\em standards conformance} cannot be guaranteed in the
canned portions of the representation.  On the other hand, a significant
amount of {\em tailoring} can still be done in the non-canned portions
of the text;
{\em consistency} between the design in the KB and the documentation text
can still largely be guaranteed; and {\em reduced-costs} may still be the
case for document creation and maintenance.  Some {\em multimodality} can
also be introduced, e.g., hypertext links can still be automatically added
to referring expressions.

Thus, hybrid action representations reduce the cost of creating a domain KB
for an NLG system, at the price of sacrificing some potential benefits
(most notably multilingual generation and guaranteed standards conformance).
Hybrid action representations are still superior to canned-text, however,
since they allow some amount of tailoring, make it easier to enforce
consistency within a document and between a document and a machine-readable
design database,
allow hypertext links to be automatically added to texts, etc.
Whether hybrid action representations are appropriate in a particular NLG
application depends on the goals of that application,
and in particular which of the potential
benefits of NLG are felt to be most important.

Some of the other intermediate techniques we
developed in \IDAS{} are described in \cite{idas:ijcai}.
The basic idea is the same as presented above;
the goal of intermediate techniques is to reduce the costs of using NLG
by sacrificing NLG benefits that are not regarded as important in the current
application.

\begin{figure*}
\vspace{1in}
\caption{Bilingual System}
\label{example-bilingual}
\end{figure*}

\subsection{Multilingual IDAS}
\label{multilingual idas}

Ilona Bellos, an MSc student, built a variant of \IDAS{} that could
produce output in both French and English \cite{bellos:msc},
if no hybrid action representations were used in the knowledge
base (Section~\ref{intermediate techniques}).
A screen dump of the output
of her system is shown in Figure~\ref{example-bilingual}.  This shows
three \IDAS{} responses (in a knowledge base documenting Renault cars
instead of ATEs) in both French and English; the French responses are
above the corresponding English responses.  The user switches between
languages simply by changing the user-expertise model; this loads in an
appropriate lexicon and grammar, and also sets some flags for the sentence
planner.  Colin Dick, another MSc student, worked on a Turkish version
of \IDAS{}; and Dan Cristea, a visitor from
Romania, built a morphology module for Romanian within the \IDAS{} framework
\cite{cristea:project}.

Such a multilingual adaptation of an NLG system is not unusual;
as R{\"o}sner points out \cite{rosner:multilingual-panel}, it has been common
practice since NLG research began for generators to be adapted to produce
output in multiple languages.

\section{Evaluation}
\label{eval}

\subsection{User Trials}
\label{user trials}

We were not able to perform any evaluation of \IDAS{} using the ATE
knowledge base, for various reasons. We were, however, able to perform
some user-effectiveness trials with another knowledge base that we built,
which described a racing bicycle; the results of this evaluation are reported
in this section.  Only a small number of people (3) were tested in the trials,
so the results should be considered as suggestive rather than statistically
significant.

\subsubsection{The Experiment}
\label{bike kb}

Three subjects, none of whom had much previous knowledge of bicycles,
were asked to carry out the evaluation exercise.  The exercise had three
parts:
\begin{enumerate}

\item
Subjects were given instructions on how to use \IDAS{}
and shown how to navigate around the question space.

\item
Subjects were asked to answer 15 questions about the bicycle, using
information obtained from \IDAS{}.  Example questions include:
\begin{itemize}
\item
What is the cost of the front brake cable?
\item
Imagine that you are selling this bicycle to someone who doesn't know
how to use the gear levers. Use \IDAS{} to find out how this is done, and
then explain this in your own words to the customer.
\item
True or false:
the front wheel has fewer spokes than the back wheel?
\end{itemize}
Subjects were timed, and all queries they issued to \IDAS{} were recorded.

\item
Subjects were asked to complete a questionnaire that asked for both general
opinions about \IDAS{} and specific suggestions for how it might be improved.

\end{enumerate}

The bicycle knowledge base, incidentally, described about
50 components of the bicycle, and was constructed (by hand)
in about two weeks
by a person who was familiar with \IDAS{} but had not previously constructed
any \IDAS{} knowledge bases.

\subsubsection{Analysis of Subjects' Performance}

The general result of the evaluation exercise was encouraging;
out of 45 responses in all, there were only two mistakes.  One involved
a misinterpretation of the phrase
\lingform{It is a Cinelli Super Record saddle}; the subject thought that
\lingform{Cinelli Super Record} was the name of the manufacturer, whereas in
fact \lingform{Cinelli} was the name of the manufacturer while
\lingform{Super Record} was the saddle's model name.  The other mistake
involved an incorrect description of how the gear levers worked;
the relevant information was in this case being communicated unambiguously
by \IDAS{}, but it probably would have been easier to understand if
accompanied by a diagram.

It was also encouraging that users managed to navigate through \IDAS{}'s
\scare{hyperspace} very efficiently, despite not having much experience
with the system.  Of the 132 queries issued to \IDAS{} by the users,
\begin{itemize}
\item 57 (43\%) conveyed information needed to respond to a question.
\item 29 (22\%) were intermediate nodes that a user had to pass through
in order to get to an information-presenting node.
\item 27 (20\%) were unnecessary and did not contribute to responding
to a question.
\item 19 (15\%) were repeats of a previous query.
\end{itemize}
Subjects were not asked to attempt to minimize the number of
queries, and some of the \scare{unnecessary} queries
were in fact due to subjects randomly browsing through the
knowledge base.  An analysis of the remaining unnecessary queries suggests
that many were due to subjects being unfamiliar with \IDAS{} in general and
the bicycle knowledge base in particular; experienced \IDAS{} users would
presumably be more efficient in their use of the system.

Subjects in some cases went
down a wrong path in hyperspace when attempting to get information,
but in all cases managed to quickly recover from this.
Subjects were also able to combine information from several \IDAS{} queries
in a single multisentence response; this supports the claim
(see \cite{levine:nlg-workshop,idas:applied-acl}, and Section~\ref{operations}
of this paper) that it is sufficient for \IDAS{} to generate short
responses, and rely on the user to be able to put them together as necessary.

In summary, there is clearly room for improvement in both the way \IDAS{}
uses text to present information, and in its use of hypertext mechanisms.
Nevertheless, the system's performance seems to be quite reasonable.
The fact that subjects answered 95\%  of the questions correctly suggests that
in the great majority of cases \IDAS{} is presenting information in a clear
and accessible manner, and the fact that only one-third of the queries
were unnecessary indicates that subjects in most cases managed to
navigate around \IDAS{}'s hyperspace without excessive difficulty.

\subsubsection{Subject's Comments}

After completing the exercise, subjects were asked for comments and suggestions
about \IDAS{}.  In general, the comments were quite favorable and supportive.
More specifically,

\begin{itemize}
\item
There were several complaints about the details of the \IDAS{} hypertext
interface (use of mouse buttons, positioning of windows, etc).  These problems
could easily be fixed by building a better user interface.
\item
Some subjects wanted to be able to ask more questions
(e.g., \lingform{how does it work?}).
\item
Subjects commented that \IDAS{}'s texts were very concise,
but in general agreed
that this was appropriate in the context of helping users perform specific
tasks (as opposed to teaching them general information about a bicycle).
\item
Subjects felt that finding information by searching through \IDAS{}'s
question-space (hyperspace) was quicker and easier than
finding it in paper documentation.
\item
Subjects commented that having some graphics output (in particular,
a picture of the bicycle) would have been useful (the version of \IDAS{}
used in the evaluation trial was not able to display diagrams of any kind).
\end{itemize}

\subsection{Industrial Reaction}

In addition to the quantitative user-evaluation trials,
we also solicited informal
reactions and comments from our industrial collaborators, and from
interested people in other industrial R\&D establishments.
Although this is not as
rigorous as the data from our user-evaluation trials, it is valuable in
helping to answer broader questions, including in particular what potential
benefits of NL generation are most likely to be useful in the real world.

These reactions can perhaps best be summarized by going over the benefits
described in Section~\ref{benefits}.
The following comments are reconstructed from many comments made by many
people over the course of many meetings and demonstrations;
we are not claiming that they represent anyone's opinion except our own.

\subsubsection{Reduced Cost}

In retrospect, we underestimated the cost of building a knowledge base
that can support NLG.
This is not a cheap endeavour, and it may be unrealistic to hope that
its cost will be less than the cost of simply writing documentation directly.
Even if some information can be extracted from an existing database or
knowledge base (e.g., a CAD system), additional information will almost certainly
be required for NL generation, and entering it will not be cheap.
No existing CAD system that we are aware of, for example, includes the kind of
design rationale information that an NLG system would need in order to be
able to respond to a What-is-its-purpose question (which was one of \IDAS{}'s
basic questions).

It may, however, be more realistic to expect that the NLG
approach can reduce the cost of document maintenance, even if it does not
reduce the cost of initial document creation; and document maintenance
can be a larger proportion of total life-cycle cost than initial
document production.  By document maintenance, we primarily mean the cost
of updating documentation when the hardware being documented changes, not
the cost of fixing spelling and grammatical errors
(Section~\ref{reduced cost}).
Once the initial machine
design has been entered into the CAD or KR systems, many of the most common
changes to that design (e.g., a new configuration, or an upgraded component)
can be made fairly easily, and in a manner that can be well supported by
authoring tools such as the one developed by our industrial collaborators
(Section~\ref{knowledge base}); making changes in
this manner and then regenerating the documentation may well be cheaper than
revising the documentation by hand.

Many of the enhanced maintainability advantages of
the \IDAS{} NLG system, however, were also present in the
object-oriented canned-text system,
which also supported creating new configurations and upgrading
components in existing configurations.
The maintainability advantages
of the NLG version of \IDAS{} over the canned-text system may thus
not be that high, at least in the ATE domain.  So, while
maintainability is extremely important and should be kept in mind
for all documentation systems, it is hard to claim that it is a particular
benefit of NLG systems; most of the benefits we observed could be obtained
simply by building an \scare{object-oriented} knowledge base
that can represent Is-A and Part-Of hierarchies of components,
and then associating canned texts with the objects represented in this KB.

Perhaps a more promising way of justifying the expense of creating
a knowledge base is to ensure that the design knowledge it contains
is used in many ways, not just for document generation.
If knowledge base authoring is thought of merely as a replacement for
document authoring, then indeed it might seem to require unreasonable
resources. But if the \lingform{knowledge base + NLG}
architecture is presented as a
solution to a wider need to design, reason about, and present products,
and if this can be integrated with the normal product development process,
then it looks much more attractive. In a follow-on project
to \IDAS{} \cite{levine:nlg94},
we are hoping to evolve our authoring tool into a general requirements capture
tool to be used by engineers right from the start of the design process. In
such a situation we hope that the cost of a small amount of extra authoring
(largely collaborative) will be amply repaid by advantages gained by a number
of sections of the company.

Another point that was raised in our discussions was that it was desirable
to have a single tool capture both the \scare{normal} design
information, and the \scare{extra} information
needed for NLG or other knowledge-based processing.
This may require the design engineer to do more work than if he or she
just enters \scare{design} information, and someone else enters
\scare{documentation-related} information, but the total amount of
effort will be less with an integrated tool, and there will be far fewer
opportunities for inconsistencies.  Also, if the knowledge base is being
used to support many kinds of reasoning, it may be hard in any case to make
a clear distinction between \scare{design} and \scare{extra} information.

In summary, we would now be cautious about claiming that generating
documentation from a knowledge base will reduce direct document-creation costs
if a special knowledge-base has to be created for the NLG system.
Cost-reduction is perhaps only likely if most of the information needed for
NLG can be extracted from information that is being used for other purposes;
and cost-reduction may be more likely for document
maintenance than initial document creation.

\subsubsection{Consistency}

Ensuring that a document is in fact consistent with a design is a very
important benefit to industry,
and one that we did not fully appreciate when we started
\IDAS{}.  It is difficult to ensure that the design described in a
human-readable text document is the same as the design described in a
machine-readable design database, and this problem becomes especially severe
when a document is being updated (e.g., to reflect changes in the hardware),
and the document updater is not the original document author.
Furthermore, inaccuracies in
documentation are very worrisome to companies, because they can cause
customers to become annoyed and consider switching to another supplier,
and because they may result
in a company being legally liable if customers misuse a product.
Our discussions suggested that many companies might be willing to accept
higher costs for document production if the resultant documents had fewer
inaccuracies, and that increased document accuracy is in fact
one of the most important potential benefits of using NLG to produce
technical documentation. Indeed, for some applications, accuracy is much more
important than the quality of the text.

\subsubsection{Guaranteed Conformance to Standards}

Ensuring that a document conforms to relevant standards is another
important potential benefit of NLG that we did not initially appreciate,
and that proved to be extremely important to many of the industrial people
we talked to.  Writing standards in particular can seem unnatural to
human authors (e.g., AECMA Simplified English \cite{simplified-english:book}
prohibits \lingform{Test the power supply} and requires
\lingform{Do a test on the power supply} instead), and training authors
to obey the standards can be a non-trivial task.  With NLG systems, however,
the relevant standard can simply be incorporated into the system's
grammar, lexicon, and planning rules, and then all output will be guaranteed
to meet the standard.  Indeed, it is
probably easier to generate Simplified English than full English, because
many of the syntactic, lexical, and other choices that have to be made when
generating full English are already specified in the Simplified English
standard, and hence the NLG system does not have to worry about them.

One thing that was clear from talking to our industrial contacts, incidentally,
is that no one wanted systems that produced
linguistically complex output.  All potential users we talked to
preferred to have technical documentation presented as simply as possible;
the use of complex syntactic or lexical constructions, which has been the
focus of much academic research, was a minus, not a plus, as far as these
people were concerned.

\subsubsection{Multilinguality}

Producing documents in several languages from a single domain KB is certainly
technically possible \cite{inlgw92:multilingual-panel}, and indeed a
bilingual French/English version of \IDAS{} was built by one of our
MSc students (Section~\ref{multilingual idas}).  Perhaps the main disadvantage
of multilingual generation (in addition to the need to create multiple
lexicons and grammars) is that it disallows the use of hybrid action
representations and similar otherwise-useful intermediate techniques
(Section~\ref{intermediate techniques}).
All knowledge to be communicated must be properly encoded in the underlying
deep representation, and this can
make the domain KB authoring task more difficult.

The level of interest in multilingual generation varied greatly among
our industrial contacts.  Some people (especially those working for firms
that produced consumer goods) thought this was potentially very valuable.
People working for aerospace and other heavy industrial firms, however,
often felt
that a better way to reach international customers was to produce documents in
Simplified English \cite{simplified-english:book} and similar English dialects
that are designed to be easily readable by non-native speakers.
The cost of generating
documents in multiple languages is not zero, after all, even if a proper
knowledge base exists, because it will probably still be necessary for
human editors and quality assurance personnel
to check the translated documents before they get sent out to customers.

There are also cases
where multilingual output is required by law.  This indeed is part
of the justification of the FoG weather-report generation system
\cite{goldberg:ieee94}
(weather reports in Canada must be distributed in both French and English).

\subsubsection{Tailoring}

One of the initial goals of \IDAS{} was to be able to tailor its output to
different kinds of users, including operators, maintenance
engineers, and programmers.  There has been a substantial amount of research in
user-tailoring in the NLG community, including for example
\cite{paris:cl,mccoy:cl,uc:cl,eurohelp:book}.
Following this
research and incorporating some ideas of our own, we built into \IDAS{}
separate user-task, user-expertise, and discourse models; as a result,
in some cases perhaps 50 different responses could potentially be produced
for one query, depending on the value of these contextual parameters.

Unfortunately, it turned out that such a high degree of variability was
{\em not} desirable for our industrial collaborators, because it made
QA (quality assurance) more difficult and expensive.  All responses generated
by our system would need to be examined by the QA department before our
system could be sent to customers, and having 50 variants of a response
made that task 50 times more difficult.  A small number of variations was
thus perhaps useful, but utilizing a rich fine-grained contextual model
to produce many response variants was definitely {\em not} desirable.

We also observed that \IDAS{} users often
used the hypertext followup mechanism to clarify terms or actions they
did not understand; they simply clicked on unfamiliar words or actions,
and in most cases got sufficient information from the followup
text to enable them to continue with their original task (even though \IDAS{}
hypertext followups were not originally designed or intended to serve
as a glossary or term-explanation mechanism).
Many commercial on-line help systems of course use hypertext in this
way; the user clicks on a word
he or she doesn't understand, and a glossary entry or new help window
appears.  The hypertext approach both gives the user more direct control
over what he or she sees, and also avoids the QA costs of the text-modifying
tailoring that we performed in \IDAS{}.  In many applications, hypertext
mechanisms may turn out to be the most appropriate technique for supporting
users with different tasks and expertise levels.

One final point is that most of the other people from industry whom we
talked to (besides our direct
industrial collaborators) did not seem very interested in
tailoring responses, perhaps because they were more interested in cutting
the life-cycle costs of documentation (e.g., including maintenance, translation,
and editing for standards conformance) than in improving documentation
effectiveness.

\subsubsection{Multimodality}
\label{multimodality}

Both our discussions and our user-effectiveness trials emphasized
that any useful technical documentation generation system {\em must} be able
to produce output that includes visual formatting, hypertext links, and
diagrams whenever appropriate (and when allowed by the medium).
A system that generates
\scare{technical documentation} that consists only of a sequence of words and
sentences may be an interesting academic exercise, but it is unlikely to be
useful in real applications.

\IDAS{} did produce hypertext; this was one of its
original design goals.  Indeed, our experience has been that if one is
going to all the trouble to generate NL text from a knowledge base,
adding hypertext followup links to this text is a relatively low-cost
increment to the basic NLG system \cite{idas:applied-acl}.
\IDAS{} can also use some visual formatting devices;
this capability is not as extensive as it should be,
but the proper use of such formatting devices is an under-researched
area in NLG.

The Edinburgh \IDAS{} system is not able to perform any kind of graphic
generation, although one version of the system can display canned bitmaps
in response to certain queries.  This is a definite weakness of the system,
and an automatic technical-documentation
generation system that is used in real applications may need
to possess more sophisticated graphic abilities.  There has been some
research on combining text and graphics generation
(e.g., \cite{feiner-mckeown:aaai90,wip:aij}),
but this work has tended to stress very
\scare{principled} ways of doing this,
which may be too costly (in terms of both the amount of domain knowledge
and the amount of compute time required) to be practical in realistically
sized systems; further research probably needs to be done on \scare{cheaper}
ways of combining text and graphics generation.

\section{Lessons Learned}
\label{lessons}

Perhaps the most important lessons we have learned from \IDAS{} are:

\begin{itemize}

\item
Document production should as much as possible utilize information in
existing design (and other) databases; if more information is needed,
it should ideally be useful for other purposes in addition to document
generation (e.g., consistency checks).  The capture of the necessary information,
and the production of documents, should be an integral part of the design
environment.
\item
The output text should be kept as linguistically simple as possible, with
relevant writing and content standards being followed.  Graphical mechanisms
(including visual formatting and hypertext input as well as diagram
production) should be used when they are appropriate.
\item
Automatic document generation is probably best justified
in terms of guaranteeing
consistency of documents with the actual designs, guaranteeing that
relevant standards are followed, and simplifying the process of updating
documents to reflect changes in the documented hardware.
It may be more difficult to justify automatic
document generation on the basis of
reducing the costs of initially creating a document.
\item
Multimodal techniques (such as automatic insertion of hypertext links)
hold promise as a way of increasing the effectiveness of generated
documentation; user-tailoring may be less promising, unless a way can be
found to solve the quality-assurance problem.
\end{itemize}

In conclusion, we believe
that there is great potential in using natural-language generation
to automate the process of producing technical documentation,
if the developers of such
systems have a clear idea of the costs and benefits of NLG, and hence of
the niches in which it might most usefully be applied.
The technology is not a panacea that will instantly cut document-production
costs to zero, but when used appropriately it has great promise in reducing
the total life-cycle costs of documentation, in making documentation more
accurate and effective, and in enabling design engineers to spend more
of their time on designing and less on documenting.

\section*{Acknowledgements}

The \IDAS{}
project was partially funded by UK SERC grant GR/F/36750 and UK DTI grant
IED 4/1/1072, and we are grateful to SERC and DTI for their support of this
work.  We would especially like to thank the \IDAS{} industrial
collaborators ---
Inference Europe, Ltd.; Racal Instruments, Ltd.; and Racal Research, Ltd. ---
for all the help they have given us in performing this research.
We would also like to thank the many interested people from other
industrial organizations (including Andersen Consulting, British Aerospace,
COWIconsult, Dassault Aviation, General Motors, and Sun Microsystems)
who have spent significant amounts of time discussing \IDAS{}
and natural-language generation with us; the first author would also like
to thank the people at CoGenTex (especially Richard Kittredge and
Tanya Korelsky)
for their comments and advice.  We are also very grateful
for the effort contributed by the
visitors and MSc students who worked on \IDAS{}, including Ilona Bellos,
Dan Cristea, Colin Dick, Sam Marshall, and Michael Pake.
It goes without saying, of course,
that all opinions expressed in this paper are our own, and we are solely
responsible for any errors or mistakes in the text.



\setlength{\baselineskip}{13pt}		
\newpage
\pagestyle{empty}
\begin{verse}
\fcomment{The initial query}\\
\fcomment{This is the `What is the lever' query in Figure~\ref{example-dump}}\\
Basic question is WHAT\\
Component is LLEVER-TEST-HEAD12\\
Task is OPERATIONS\\
User-model is SKILLED\\
Focus-list is NIL\\
~\\
\fcomment{The output of content-determination}\\
Schema function is\\
{\tt ~~(IDENTIFY-SCHEMA :BULLET? NIL :UNABBREVIATE? T)}\\
Schema properties are ((COLOUR BLACK))\\
Followups are (WHERE)\\
~\\
\fcomment{The output of sentence planning}\\
SPL is
\begin{verbatim}
(S1543 / IDENTITY
  :DOMAIN (LLEVER-TEST-HEAD12 /
           |locking lever| :PRONOUN YES)
  :RANGE (R1542 /
           |locking lever| :DETERMINER INDEFINITE
           :RELATIONS
            ((R1545 / |colour| :DOMAIN R1542
              :RANGE (R1544 / BLACK)))))
\end{verbatim}

\fcomment{The output of surface realisation}\\
It is a black locking lever.
\end{verse}

%
\vspace{1in}
{\bf Figure 2: A Trace of Response 8 in Figure 1}

\end{document}